\documentclass[]{interact}

\usepackage{epstopdf}% To incorporate .eps illustrations using PDFLaTeX, etc.
\usepackage[caption=false]{subfig}% Support for small, `sub' figures and tables
\usepackage{bm}
\usepackage{color}
\usepackage{url}
\usepackage{cite}
\usepackage{hyperref}
\usepackage{wrapfig}

\usepackage{lipsum}

\usepackage{amsmath,amsthm}

\usepackage{algorithm}
\usepackage{algpseudocode}

\theoremstyle{plain}
\newtheorem{thm}{Theorem}
\newtheorem*{thm*}{Theorem}

\usepackage{diagbox} 
\usepackage{multirow}
\usepackage{multibib}
\usepackage[normalem]{ulem}

\theoremstyle{plain}% Theorem-like structures provided by amsthm.sty

\theoremstyle{definition}

\theoremstyle{remark}

\begin{document}

\articletype{ARTICLE TEMPLATE}% Specify the article type or omit as appropriate

\title{Emergent Communication through Metropolis-Hastings Naming Game with Deep Generative Models}

\author{
\name{Tadahiro Taniguchi\textsuperscript{a}\thanks{CONTACT T. Taniguchi. Email: taniguchi@ci.ritsumei.ac.jp}, %
Yuto Yoshida\textsuperscript{a}, %
Akira Taniguchi\textsuperscript{a}, and %
Yoshinobu Hagiwara\textsuperscript{a}}
\affil{\textsuperscript{a}Ritsumeikan University \\ 1-1-1 Noji\ Higashi, Kusatsu, Shiga 525-8577, Japan.}
}

\maketitle

\begin{abstract}
Constructive studies on symbol emergence systems seek to investigate computational models that can better explain human language evolution, the creation of symbol systems, and the construction of internal representations.
Specifically, emergent communication aims to formulate a computational model that enables agents to build efficient sign systems and internal representations.
This study provides a new model for emergent communication, which is based on a probabilistic generative model (PGM) instead of a discriminative model based on deep reinforcement learning.
We define the Metropolis-Hastings (MH) naming game by generalizing previously proposed models. It is not a referential game with explicit feedback, as assumed by many emergent communication studies. Instead, it is a game based on joint attention without explicit feedback.
Mathematically, the MH naming game is proved to be a type of MH algorithm for an integrative PGM that combines two agents that play the naming game.
From this viewpoint, symbol emergence is regarded as {\it decentralized Bayesian inference}, and semiotic communication is regarded as inter-personal cross-modal inference. This notion leads to the {\it collective predictive coding hypothesis} regarding language evolution and, in general, the emergence of symbols.
We also propose the inter-Gaussian mixture model (GMM)+ variational autoencoder (VAE), a deep generative model for emergent communication based on the MH naming game. In this model, two agents create internal representations and categories and share signs (i.e., names of objects) from raw visual images observed from different viewpoints.
The model has been validated on MNIST and Fruits 360 datasets. Experimental findings demonstrate that categories are formed from real images observed by agents, and signs are correctly shared across agents by successfully utilizing both of the observations of agents via the MH naming game. Furthermore, scholars verified that visual images were recalled from signs uttered by agents. Notably, emergent communication without supervision and reward feedback improved the performance of the unsupervised representation learning of agents.
\end{abstract}


\section{Introduction}
Constructive studies on symbol emergence systems, which are multi-agent systems that can make symbols or language emerge and use them for communication, are crucial for understanding human language and cognition and creating robots that can adapt to our semiotic communication~\cite{Steels97,deacon1998symbolic,taniguchi2018symbol,taniguchi2016symbol,taniguchi2021semiotically}.
Specifically, emergent communication aims to build a computational model that enables agents to build efficient sign systems and internal representations.
%In addition, obtaining a reasonable computational model for emergent communication is important for the realization of more semiotically adaptive human-robot collaboration.
Language (and symbol systems in general) features a dynamic nature. It changes dynamically through time in terms of forms and meanings.
Despite the time-varying properties, symbol systems consistently enable individuals to communicate information about external objects. In other words, certain cognitive and social dynamics in multi-agent systems enable agents to form a symbol system in an emergent manner and offer a function of semiotic communication to the agents. The system is considered a complex system with an emergent property and referred to as {\it symbol emergence system}~\cite{taniguchi2016symbol}.
Importantly, such emerged symbols assist agents not only in communicating via signs but also in appropriately categorizing objects. A proper reciprocal reliance between sign sharing and perceptual category formation, including representation learning, is critical for the computational model of {\it perceptual symbol systems}~\cite{barsalou99}.

%There are various requirements for the computational model of emergent communication.
The significant challenge in semiotic communication, that is, communication using signs, is that agents can neither inspect each other’s brain states nor directly transmit meanings~\cite{steels2015talking}.
A {\it symbol} is a triadic relationship of a {\it sign}, an {\it object}, and an {\it interpretant}following the terminology of Peircian semiotics~\cite{Chandler2002}.
The challenge in emergent communication is developing not only a model that enables an artificial agent to form symbol systems for better communication and cooperation but also one that can explain language acquisition and symbol emergence in humans. This notion has been a long-term challenge in {\it symbol emergence in robotics}~\cite{taniguchi2016symbol}.

Scholars examined language games, such as naming and referential games with explicit feedback, for modeling emergent communication over time. Many studies in this field were based on variants of the Lewis signaling game~\cite{lewis2008convention}. Steels and related scholars in artificial life and developmental robotics conducted a wide range of pioneering works as a synthetic approach to language evolution~\cite{AIBO, steels2015talking,Steels95, Steels05,Steels08,Spranger11, Spranger12,Spranger14,Vogt02,Vogt05,Beule06,Bleys15}. %These studies were conducted not only in simulation but also in a real-world setting using robots with an emphasis on embodied interactions.
Following the publication of key works by Foerster et al. and Lazaridou et al. \cite{Foerster,Lazaridou17}, studies on emergent communication have been revived. Many studies have been conducted~\cite{Graesser19,Chaabouni21,lazaridou2020emergent,Tucker21,Mu21,jiang2018learning,kim2020communication,li2019ease,eccles2019biases} due to the invention of deep reinforcement learning. The reason is that the representation-learning power of deep learning is required to realize symbol emergence based on raw sensory data. These models can be regarded as emergent communication models based on {\it discriminative models} according to machine learning perspectives.

However, as Tomasello importantly pointed out, this type of pointing-and-naming game with explicit feedback is not representative of the vast majority of word-learning situations that children encounter in daily life~\cite{tomasello2005constructing}. Therefore, language games with explicit rewards or supervisory feedback are not suitable models from the developmental point of view.
In contrast, it is widely known that a human infant holds the capability for {\it joint attention} during the early developmental stage, and the skill becomes the foundation of language acquisition~\cite{cangelosi2015developmental}.
In other words, the assumption of joint attention is more plausible than the assumption of explicit feedback in a language game from the developmental perspective.

At the same time, {\it generative models} are widely used for modeling representation learning and concept formation based on multimodal sensory information~\cite{kingma2013auto,suzuki2016joint,taniguchi2020improved,Nakamura09,Nakamura2014}. In cognitive science and neuroscience, the generative perspective of cognition, which is also known as the {\it free-energy principle} and {\it predictive coding}, has become dominant as a general principle of cognition~\cite{friston2010free,hohwy2013predictive,ciria2021predictive}. The {\it world model}-based approach to artificial intelligence also follows this view~\cite{FRISTON2021573,planet_b}.

\begin{figure}[tb!p]
    \centering
    \includegraphics[width=0.6\linewidth]{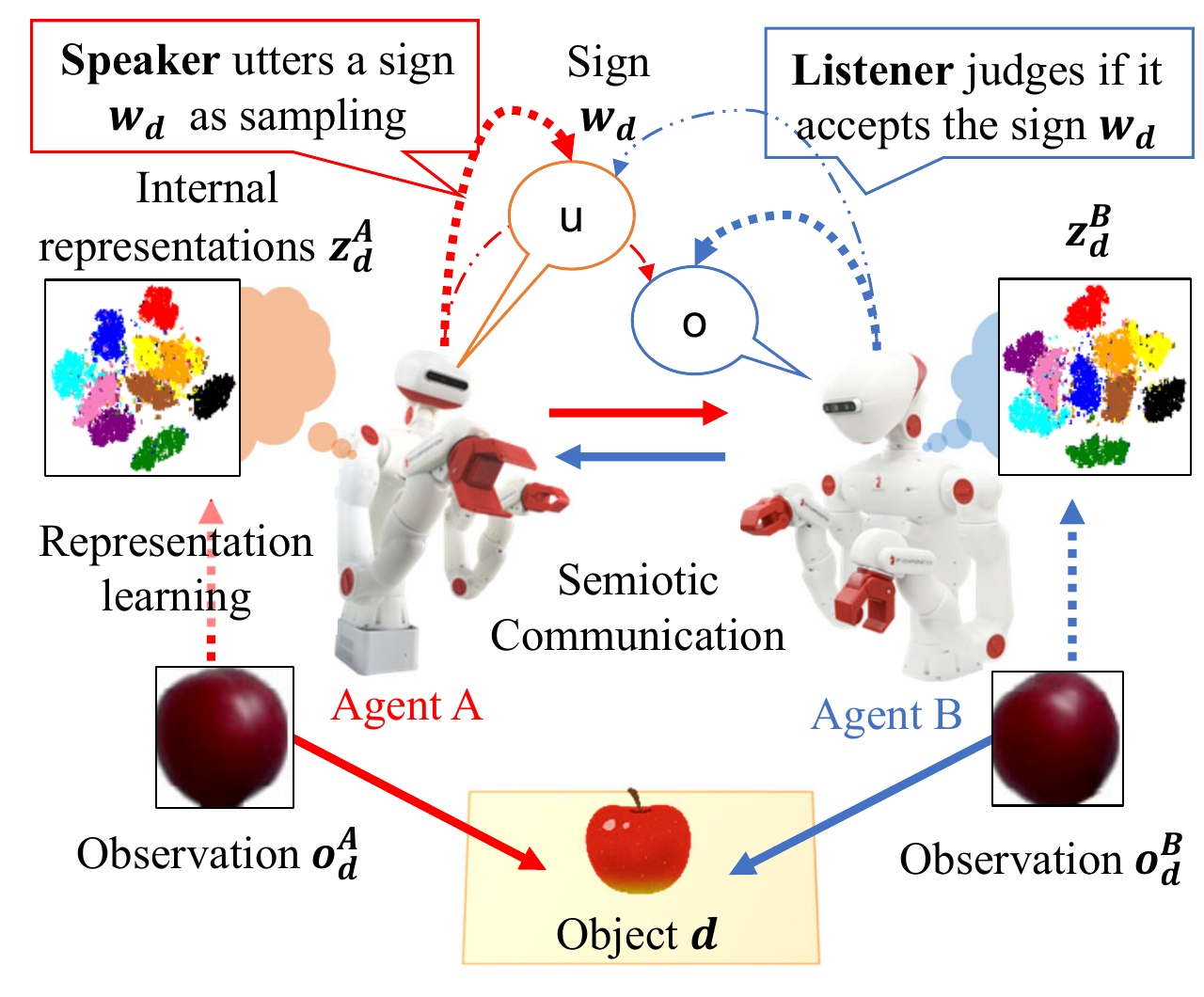}
    \caption{Overview of MH Naming Game}
    \label{fig:overview}%\vspace{-5mm}
\end{figure}

With this context in mind, this study presents a novel emergent communication framework based on deep probabilistic generative models (PGMs). We first define the Metropolis-Hastings (MH) naming game. Hagiwara et al.~\cite{hagiwara2019symbol} initially introduced this type of game for a specific probabilistic model. In the current study, the MH naming game is generalized and formally defined by generalizing the idea. The game appears to be similar to the original naming game; however, it is not. The MH naming game does not require any explicit feedback between agents but assumes the existence of joint attention inspired by developmental studies.
The MH naming game is completely based on PGMs and is mathematically demonstrated to be the same as the Metropolis-Hastings algorithm for the model. The model represents the generative process of the representation-learning and sign-sharing processes of two agents as a whole. The emergent communication is regarded as {\it decentralized Bayesian inference} (see Theorem 1). Figure~\ref{fig:overview} provides an overview of the MH naming game. Semiotic communication is defined as {\it inter-personal cross-modal inference} when a speaker provides the name of a target object, and a listener recalls the picture of the item from the name.

The limitation of the models proposed by Hagiwara et al.~\cite{hagiwara2019symbol,hagiwara2022multiagent} is that they do not involve deep generative models and cannot enable agents to conduct symbol emergence on raw images and to image ( i.e., reconstruct) objects corresponding to signs. They also did not provide a general theory for the MH naming game. To address these aspects, the current study presents an inter-Gaussian mixture model (GMM)+ variational auto encoder (VAE) or inter-GMM+VAE, a deep PGM, and an inference procedure for the model. The inference procedure is based on an MH naming game and a decomposition-and-communication strategy for modeling emergent communication based on deep probabilistic models~\cite{taniguchi2020neuro}.

The main contributions of this paper are twofold.
\begin{itemize}
    \item By generalizing earlier studies, we establish the MH naming game to provide an emergent communication framework based on PGMs. In contrast to conventional language games, it assumes joint attention instead of explicit feedback between agents. We demonstrate that, in this framework, emergent communication is equal to the MH algorithm of a PGM that represents two agents. In other words, emergent communication is formulated as external and internal representation learning based on the decentralized Bayesian inference.
    \item We propose inter-GMM+VAE and its inference procedure as an MH naming game that enables two agents to undertake emergent communication, classify raw images and share signs that represent them in a cooperative manner. On two datasets, namely, MNIST and Fruits 360, we illustrate that emergent communication based on inter-GMM+VAE enables two agents to build categories and share signs at the same level as centralized inference\footnote{Code available at \url{https://github.com/is0383kk/SymbolEmergence-VAE-GMM}}.
\end{itemize}

%\begin{wrapfigure}{r}[0pt]{0.40\linewidth}\vspace{-5mm}
%\end{wrapfigure}

\section{Metropolis-Hastings Naming Game}\label{sec:MH-NG}
The MH naming game is a form of language game played between two agents (Figure~\ref{fig:overview}). In general, the game is played as follows. An agent views an object and tells the name based on its percept, that is, the internal state inferred from its observations. The agent says a word (i.e., a sign) corresponding to the object in a probabilistic manner (i.e., sampling a word from the posterior distribution over words). A counterpart, that is, a listener, determines whether or not it accepts the word based on its belief state. Afterward, they alternate their roles or take turns. This process does not involve explicit feedback from the listener to the speaker. In contrast, we assume joint attention, where the listener knows which object the speaker is looking at. In this section, we depict that the MH naming game can be derived as an approximate Bayesian inference procedure of a certain PGM that represents two agents as an integrative system.

\begin{figure}[tb!p]
    \centering
    \includegraphics[width=1.0\linewidth]{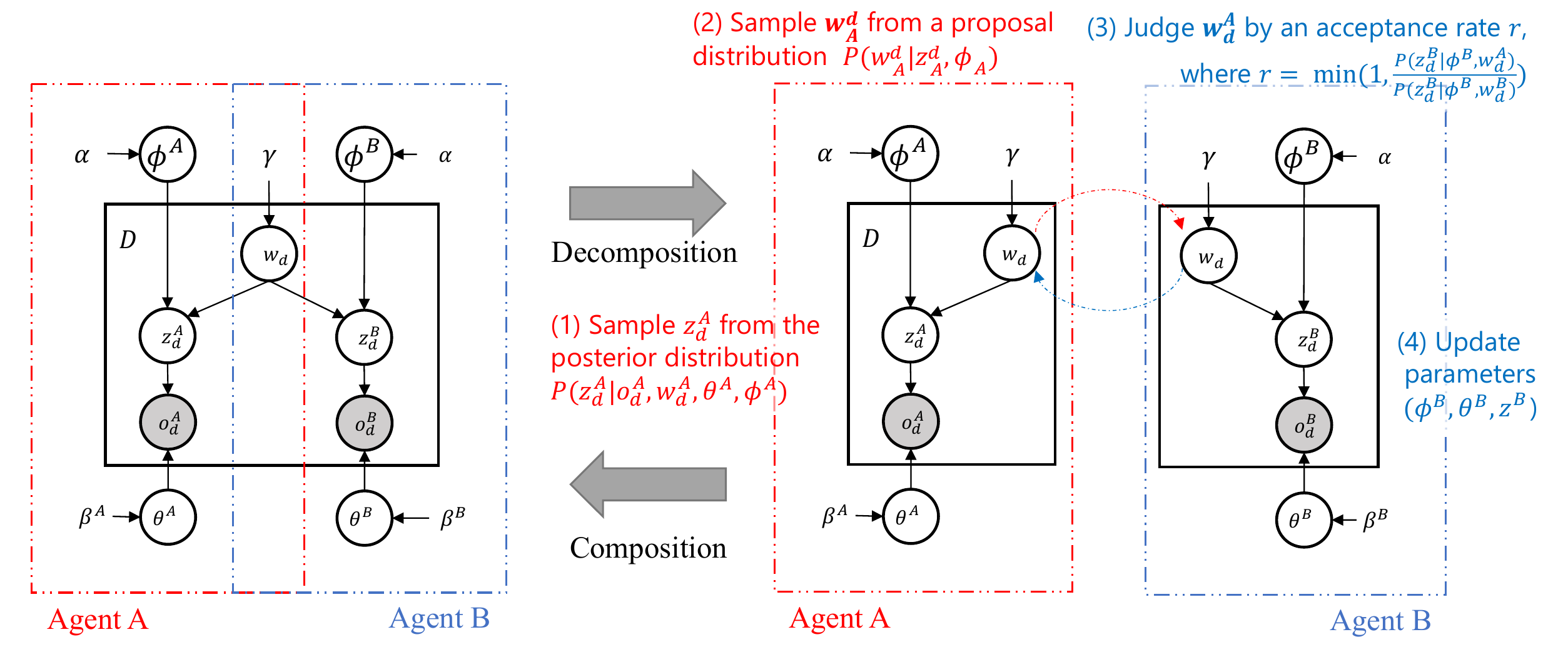}
    \caption{Probabilistic graphical model considered for the MH naming game and its decomposition. When Agent A is a speaker, Agent A samples (1) $z_d^A$ using observation $o_d$ of the $d$-th object as perception and (2) $w_d^A \sim P(w_d|z_d^A, \phi^A)$ as an utterance. The variable $w_d$ is regarded as a sign that Agent A utters. After receiving the sign $w^A_d$, (3) Agent B judges if it accepts the sign by comparing $P(z_d^B|\phi^B, w_d^A)$ and $P(z_d^B|\phi^B, w_d^B)$ in a probabilistic manner. After completing (1)-(3) about all target objects, (4) Agent B updates its internal variables by sampling. Subsequently, the agents make a turn, and Agents B and A become a speaker and a listener, respectively. The total process is revealed to be a sampler based on the MH algorithm.}
    \label{fig:gm-mhng}
\end{figure}

The left panel in Figure~\ref{fig:gm-mhng} presents a PGM that integrates two PGMs that represent two agents with a latent variable $w_d$. This notion can be regarded as a PGM of a variant of multimodal VAEs~\cite{suzuki2022survey}.
When observing the $d$-th object from a different perspective at the same time, Agent $*$ receives observations $o_d^*$ and infers internal representation $z_d^*$. Notably, $*$ represents A or B throughout this study. The graphical model shown in Figure~\ref{fig:gm-mhng} left displays a latent variable $w_d$ shared by the two agents. In the context of multimodal VAEs, $w_d$ corresponds to a latent representation that integrates two modality information, namely, visual and auditory information.
From the viewpoint of a standard inference scheme, such as the Gibbs sampling and variational inference, information about $z^A_d$ and $z^B_d$ such as posterior distributions or samples in Monte-Carlo approximation are required to update $w_d$. However, $z^A_d$ and $z^B_d$ are internal representations of Agents A and B, respectively. Therefore, each agent cannot look into the internal state of the other, which is the fundamental assumption of human semiotic communication. Metaphorically, if the brains of the two agents were connected, $w_d$ would be an internal state of the connected brain and can be inferred by referring to the internal representations $z^A_d$ and $z^B_d$ of each agent. However, it is not the case.
The question is whether or not agents can infer the shared variable $w_d$ without connecting their brains, that is, without simultaneously referring to $z^A_d$ and $z^B_d$. Thus, playing the MH naming game is the solution.

Let us decompose the generative model into two parts following the symbol emergence in robotics toolkit (SERKET) framework (Figure~\ref{fig:gm-mhng} right)\cite{taniguchi2020neuro,nakamura2018serket}.
SERKET is a framework that enables the decomposition of a PGM into several modules and derives an inference procedure in a systematic manner. A total inference procedure can be divided into inter-module communication and intra-module inference, which is the same as the usual inference procedure of the elemental module~\cite{taniguchi2020neuro,nakamura2018serket}.

The graphical models corresponding to Agents A and B are structurally the same as PGMs for representation learning, such as VAEs. $z_d^*$ is an internal representation of $o_d^*$, which is inferred under the influence of the prior distribution, which has a variable $w^*_d$. Here, lists $o^*=(o^*_d)_{d\le D}, w^*=(w^*_d)_{d\le D}, w=(w_d)_{d\le D}$, and $ z^*=(z^*_d)_{d\le D}$ are defined, where $D$ is the number of objects. In addition, $o=\{o^A, o^B \}, z=\{z^A, z^B \}, \theta =\{\theta^A, \theta^B \}$, and $\phi=\{\phi^A, \phi^B \}$.

Let us regard the sampling process $w^*_d \sim P(w^*_d|z^*_d,\phi^*)$ as the {\it utterance} of a sign $w^*_d$.
With this metaphorical assumption, the sampling of $w^*_d$ can be regarded as a naming behavior for the object $d$ by $*$.
Notably, $w^*_d$ does not mean a latent variable for Agent $*$, but a tentative sample for $w_d$ drawn by Agent $*$.
The sign can be a word, a sentence, or even an image. With this assumption, the MH naming game is defined as follows.

%Here, we assume the signs $w_d^A$ and $w_d^B$ are shared by the two agents and represented as a single node $w_d$ in the graphical model.

%   \begin{wrapfigure}{r}{0.5\textwidth} \vspace{-2\baselineskip}
%     \begin{minipage}{0.5\textwidth}
 \begin{algorithm}[tb]
  \caption{Metropolis-Hastings Communication}
 \label{alg:MH-C}
 \begin{algorithmic}[1]
 \Procedure{MH-communication}{$z^{Sp}, \phi^{Sp}, z^{Li}, \phi^{Li}, w_d^{Li}$}
 \State $w_{d}^{Sp} \sim P(w_{d}^{Sp}|z_{d}^{Sp},\phi^{Sp})$
 \State $r = {\rm min}\left(1,
    \frac{
    P(z_{d}^{Li}|{\phi}^{Li},w_{d}^{Sp})}
    {
    P(z_{d}^{Li}|{\phi}^{Li},w_{d}^{Li})         
    }
    \right)$
 \State $u \sim {\rm Unif}(0,1)$
   \If {$u\leq r$}
  \State {\bf return} $w_d^{Sp}$
  \Else
  \State {\bf return} $w_d^{Li}$
  \EndIf
 \EndProcedure
 \end{algorithmic} 
\end{algorithm}

\begin{algorithm}[tb]
 \caption{Metropolis-Hastings Naming Game}
 \label{alg:MHNG}
 \begin{algorithmic}[1]
 \State {Initialize all parameters}
 \For {$t=1$ to $T$} 
 \State  //  Agent A talks to Agent B.%: Agent A tells the names of the objects to Agent B.} 
  \For {$d=1$ to $D$}
  \State { $w_d^B \leftarrow$ MH-communication $(z^A, \phi^A, z^B, \phi^B, w_d^B)$}
  \EndFor
 \State // Learning by Agent B 
 \State $\theta^{B} \sim P(\theta^{B}|o^B,z^B,\beta^B)$
 \State  $\phi^{B} \sim P(\phi^B|w^B, z^B, \alpha^B)$
 \State // Perception by Agent B 
 \For {$d=1$ to $D$}
 \State  $z_d^{B} \sim  P(z_d^B|o_d^B, w_d^B,\theta^B, \phi^B)$
 \EndFor
 \State  // Agent B talks to Agent A. %Agent B tells the names of the objects to Agent A.} 
  \For {$d=1$ to $D$}
  \State { $w_d^A \leftarrow$ MH-communication $(z^B, \phi^B, z^A, \phi^A, w_d^A)$}
  \EndFor
 \State // Learning by Agent A
 \State $\theta^{A} \sim P(\theta^{A}|o^A,z^A,\beta^A)$
 \State  $\phi^{A} \sim P(\phi^A|w^A, z^A, \alpha^A)$
 \State // Perception by Agent A
 \For {$d=1$ to $D$}
 \State  $z_d^{A} \sim  P(z_d^A|o_d^A, w_d^A,\theta^A, \phi^A)$
 \EndFor
  \EndFor
 \end{algorithmic} 
\end{algorithm}
%     \end{minipage}    \vspace{-2\baselineskip}
%   \end{wrapfigure}

Algorithm~\ref{alg:MHNG} describes the MH naming game for the PGM shown in Figure~\ref{fig:gm-mhng}. The game consists of the following steps.

\begin{enumerate}
    \item {\bf Perception}: Speaker and listener agents ({\it Sp} and {\it Li}) observe the $d$-th object, obtain $o^{Sp}_d$, and $o^{Li}_d$ infers their internal representation $z_d^{Sp}$ and $z_d^{Li}$, respectively.
    \item {\bf MH communication}: Speaker tells the name $w_d^{Sp}$ of the $d$-th object by sampling it from $P(w_d|z_d^{Sp}, \phi^{Sp})$. The listener determines if it accepts the naming with probability $r = {\rm min}\left(1,
    \frac{
    P(z_{d}^{Li}|{\phi}^{Li},w_{d}^{Sp})}
    {
    P(z_{d}^{Li}|{\phi}^{Li},w_{d}^{Li})         
    }
    \right)$.
    \item {\bf Learning}: After MH communication was performed for every object, the listener updates its global parameters $\theta^{Li}$ and $\phi^{Li}$.
    \item {\bf Turn-taking}: The speaker and listener alternate their roles and go back to (1). % Listener and Speaker, respectively. Go back to 1.
\end{enumerate}

In contrast to referential and original naming games~\cite{steels2015talking}, the MH naming game does not use any feedback, such as rewards or supervisory signals. Instead, it assumes {\it joint attention}, that is, the two agents know that they simultaneously look at the same object from different perspectives.

Surprisingly, the MH naming game satisfies the following property.

\begin{thm}
The MH naming game is a Metropolis-Hastings sampler of $P(w,z, \theta, \phi|o)$.
\end{thm}
\begin{proof}
Regarding $z^*_d \in z^* \in z$, $\theta^* \in \theta$, and $\phi^* \in \phi$, they are drawn from the distribution conditioned on the values of the remaining variables. As a result, they are considered drawn from the Gibbs sampler, which is a type of MH sampler. Regarding $w_d \in w$, if the proposal distribution $w_{d}^{Sp} \sim P(w_{d}^{Sp}|z_{d}^{Sp},\phi^{Sp})$ is used, then the acceptance ratio of the MH algorithm~\cite{hastings1970monte} becomes  $r = {\rm min}\left(1,
    \frac{
    P(z_{d}^{Li}|{\phi}^{Li},w_{d}^{Sp})}
    {
    P(z_{d}^{Li}|{\phi}^{Li},w_{d}^{Li})         
    }
    \right)$, where $(Sp, Li) \in \{(A, B), (B, A)\}$ (see Appendix~\ref{apdx:proof}). As a result, the MH naming game functions as a Metropolis-Hastings sampler of $P(w,z, \theta, \phi|o)$. In other words, the MH naming game is a decentralized approximate Bayesian inference algorithm.
\end{proof}

We have demonstrated that the MH naming game is a {\it decentralized approximate Bayesian inference} method of a PGM that integrates two agents into a system (Figure~\ref{fig:gm-mhng} left). The MH communication (in Algorithm~\ref{alg:MHNG}) realizes the inference of $P(w|o^A,o^B)$ without inspecting each other’s brain states.
Notably, the MH naming game naturally involves role alternation \footnote{Galke et al. found that role alternation and memory restriction are critical for human language evolution but are frequently overlooked by contemporary research that uses referential game research~\cite{Galke22}. }. Nevertheless, the MH naming game, even without role alternation, that is, one-way communication, can also become a Metropolis-Hastings sampler of $P(w,z, \theta, \phi|o)$. Particularly, turn-taking is a mathematically eliminable element.

\section{Inter-GMM+VAE}\label{sec:inter-GMM+VAE}
\subsection{Generative model}
\begin{figure}[tb!p]
    \centering
    \includegraphics[width=1.0\linewidth]{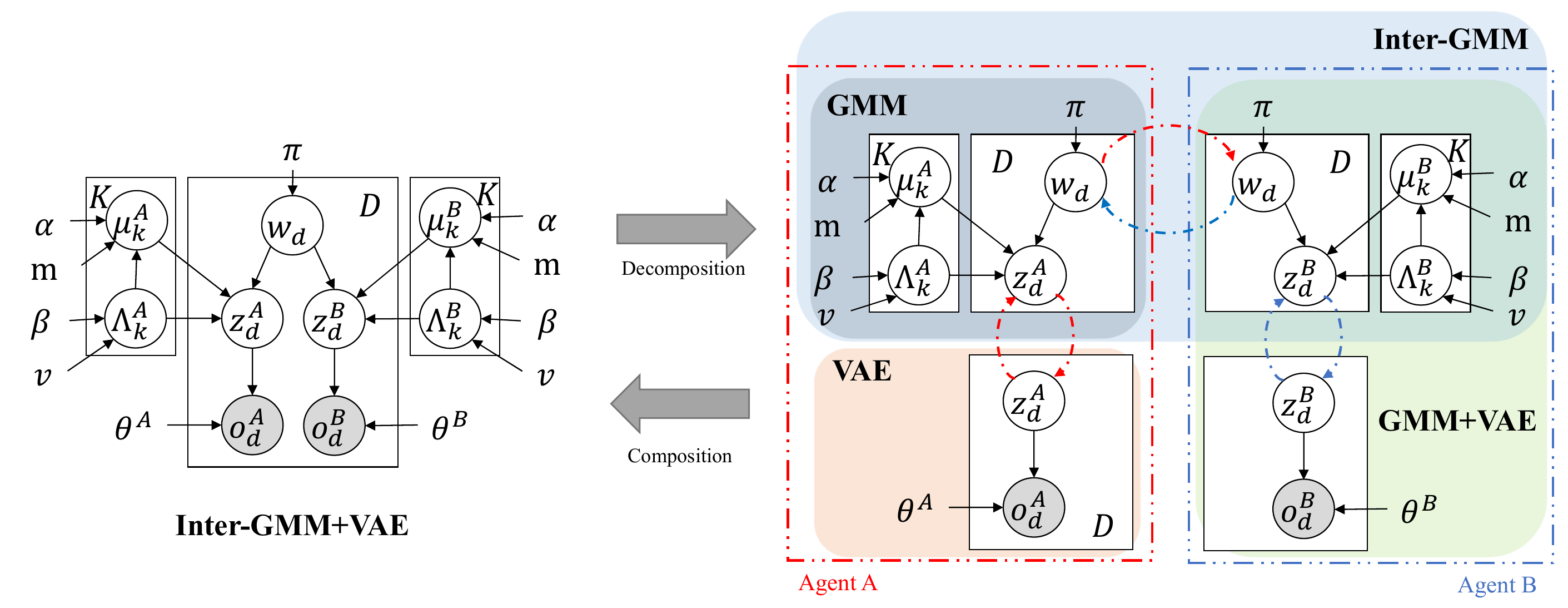}
    \caption{Probabilistic graphical model of Inter-GMM+VAE and its decomposition}
    \label{fig:inter-gmm-vae}\vspace{-5mm}
\end{figure}

We define a deep generative model for two agent-emergent communication called inter-GMM+VAE.
Figure~\ref{fig:inter-gmm-vae} illustrates a probabilistic graphical model of inter-GMM+VAE. The probabilistic generative process of inter-GMM+VAE is shown as follows.
\begin{align}
&w_d \sim {\rm Cat}(\pi) \quad  &d = 1, \ldots, D\label{eq:inter-GMM+VAE-1}\\
% &\mu_k^A, \Lambda^A_k \sim \mathcal{N}(\mu_k^A|m,(\alpha\Lambda^A_k)^{-1}){\mathcal W}(\Lambda_k^A|\nu,\beta), \ 
% \mu^B_k, \Lambda^B_k \sim \mathcal{N}(\mu_k^B|m,(\alpha\Lambda^B_k)^{-1}){\mathcal W}(\Lambda_k^B|\nu,\beta) \\
% &z^A_d \sim \mathcal{N}(z^A_d|\mu^A_{w_d},(\Lambda^A_{w_d})^{-1}), \quad
% z^B_d \sim \mathcal{N}(z^B_d|\mu^B_{w_d},(\Lambda^B_{w_d})^{-1}) \label{eq:inter-GMM+VAE-3}\\
% &o^A_d \sim p_{\theta^A}(o^A_d|z^A_d), \quad 
% o^B_d \sim p_{\theta^B}(o^B_d|z^B_d)
&\mu_k^*, \Lambda^*_k \sim \mathcal{N}(\mu_k^*|m,(\alpha\Lambda^*_k)^{-1}){\mathcal W}(\Lambda_k^*|\nu,\beta)\quad &k=1, \ldots, K\\
&z^*_d \sim \mathcal{N}(z^*_d|\mu^*_{w_d},(\Lambda^*_{w_d})^{-1})\quad &d = 1, \ldots, D\label{eq:inter-GMM+VAE-3}\\
&o^*_d \sim p_{\theta^*}(o^*_d|z^*_d)\quad  &d = 1, \ldots, D
\end{align}
where $* \in \{A, B\}$; the parameters $\mu_k^*,\Lambda_k^*$ are parameters of the $k$-th multivariate normal distributions of Agent $*$, and $\phi^* = (\mu_k^*,\Lambda_k^*)_{k\le K}$. The parameters are assumed to be generated using the normal-Wishart distribution.
The latent variable $z^*_d$ shared by the GMM and VAE components is assumed to be drawn from a multivariate normal distribution corresponding to the $k$-th sign, that is, $w_d=k$.
The discrete variable $w_d$, which represents a sign of the $d$-th object, is considered to be generated from the categorical distribution ${\rm Cat}(w_d|\pi)$.
In this research, we assume that the mixture ratio $\pi$ is a uniform distribution.
Assuming that the observations $o^*_d$ of each agent is generated from a VAE decoder $p_{\theta^*}(o^*_d|z^*_d)$ with latent variable $z^*_d$, the total generation process is described as above(Eqs. (1)-(4)).
Notably, inter-GMM+VAE can be regarded as a variant of multimodal VAEs~\cite{suzuki2022survey}.

Figure~\ref{fig:inter-gmm-vae} depicts a graphical model of inter-GMM+VAE and its composition and decomposition relationships.
Inter-GMM+VAE is obtained by composing two GMM+VAE in a manner similar to that of inter-multimodal Dirichlet mixture (MDM) is obtained by composing two MDMs in \cite{hagiwara2022multiagent}. GMM+VAE is obtained by combining GMM and VAE. Composing graphical models, particularly VAE with structured or discretized latent variables, is examined for combining the complementary characteristics of traditional PGMs, such as GMM, HMM, and LDA, with deep generative models such as VAE~\cite{johnson2016composing,oord2017neural,razavi2019generating,kuniyasurobot}. In this study, we simply combine GMM and VAE. The notation $+$, that is, the composition of two graphical models and their mutual (or simultaneous) inference, follows the convention in \cite{taniguchi2020neuro}.
We also call a generative model that consists of (\ref{eq:inter-GMM+VAE-1}) - (\ref{eq:inter-GMM+VAE-3}) inter-GMM, which is a tail-to-tail composition of two GMMs. In addition, inter-GMM+VAE can be considered a composition of inter-GMM and two VAEs.

\subsection{Inference via the MH naming game}
As explained in Section~\ref{sec:MH-NG}, the MH naming game acts as a sampling-based inference procedure of inter-GMM+VAE. However, $\theta^*$ and $z_d^*$ cannot be drawn from the analytical posterior distribution, in contrast to inter-DM and inter-MDM in~\cite{hagiwara2019symbol,hagiwara2022multiagent}, because inter-GMM+VAE involves VAE, that is, a deep generative model. 
Moreover, gradient-based optimization throughout the system cannot be employed because $w$ is assumed to be inferred through the MH naming game, that is, Markov Chain Monte Carlo (MCMC).
As a result, we use the decomposition-and-communication method employed in the (Neuro-)SERKET framework~\cite{Serket,taniguchi2020neuro}.
Mutual inference (MI) is performed between GMM and VAE. The parameters of a GMM module $(\mu_k^*,\Lambda_k^*)_{k\le K}$ are sent to a VAE module, and each VAE is trained with data-dependent prior distribution $\mathcal{N}(z^*_d|\mu^*_{w_d},(\Lambda^*_{w_d})^{-1})$. After the optimization of VAE, $z^*$ is sent to the GMM module, $\phi^*$ is inferred using Gibbs sampling, and $w$ is sampled as an utterance in the MH naming game. This MI process enables the approximate sampling of the internal variables $(z^*,\phi^*, \theta^*)$ of each agent.
Appendix~\ref{apdx:procedure} presents a diagram that depicts the overall MH naming game, that is, the inference procedure, for illustrative purposes.

\subsection{Semiotic communication as an inter-personal cross-modal inference}
Semiotic communication using sign $w_d$ (see Figure~\ref{fig:overview}) is divided into two parts. A speaker tells the name of an object $d$ by sampling $w^{Sp}_d \sim P(w_d|o^{Sp}_d)$, and a listener recalls an image of the object by sampling $o^{Li}_d \sim P(o^{Li}_d|w_d = w^{Sp}_d)$. This process pertains to an ancestral sampling across the two agents.

If we consider inter-GMM+VAE as a variant of multimodal VAE, then two observations, namely, $o_d^A$ and $o_d^B$, are regarded as multimodal observations, as previously mentioned. The cross-modal inference is the process of inferring an observation from one modality to another in multimodal generative models (e.g., $P(o_d^{Li}|o_d^{Sp})$). Furthermore, in inter-GMM+VAE, the cross-modal inference is performed across two agents. As a result, semiotic communication, in which the listener recalls an image from the sign $w_d$ provided by the speaker, is considered an {\it inter-personal cross-modal inference}~\cite{hagiwara2022multiagent}.

\section{Experiment 1: MNIST dataset}\label{sec:mnist}
\subsection{Conditions}
\begin{figure}[tb!p]
%\begin{tabular}{cc}
 % \begin{minipage}{.33\textwidth}
	\centering
    \includegraphics[width=0.4\linewidth]{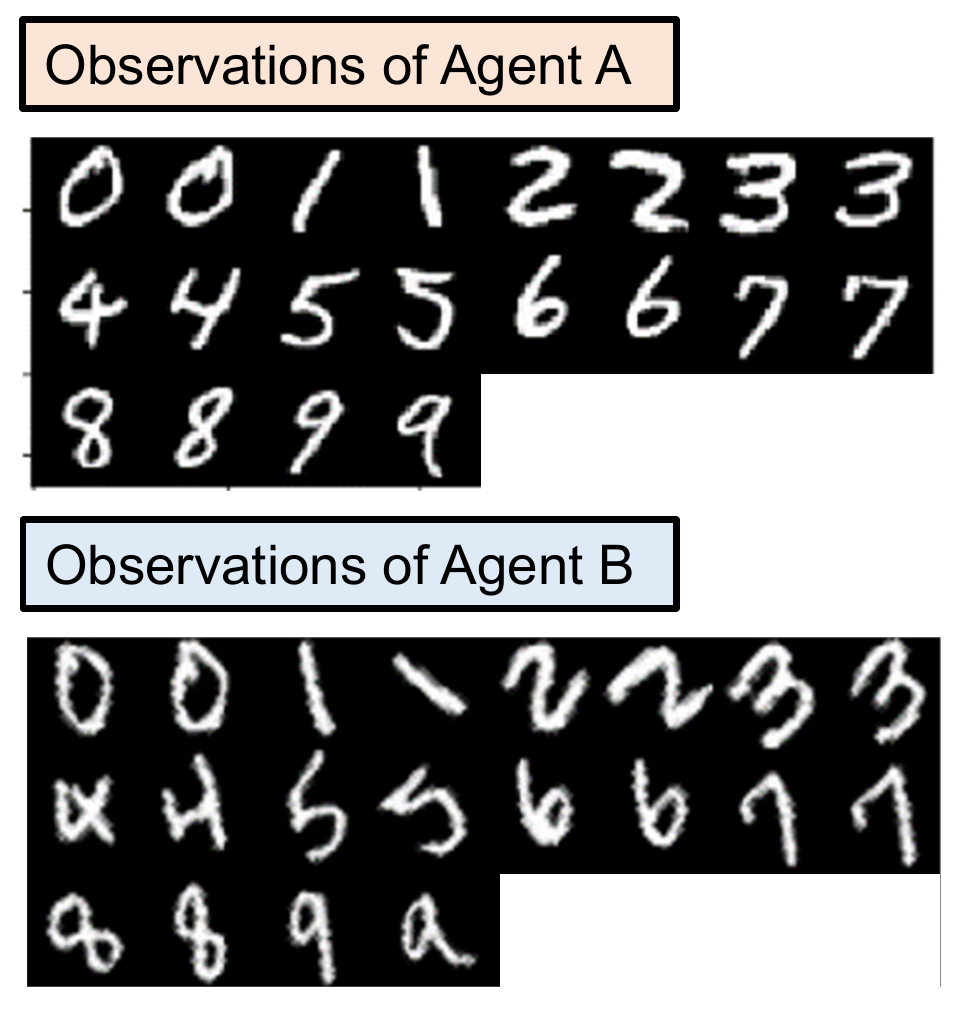}
    \caption{MNIST dataset}
    \label{fig:mnist_dataset}
 % \end{minipage}
\end{figure}

{\bf Dataset}: In this experiment, the MNIST dataset\footnote{The MNIST database: \url{http://yann.lecun.com/exdb/mnist/}} is used to validate the proposed model.
The MNIST dataset consists of $28 \times 28$ pixels handwritten character images from 0 to 9. Agents A and B are assumed to observe the same object from different perspectives.
In this experiment, we used raw MNIST data for the observations of Agent A, and MNIST data rotated 45° to the left for observations of Agent B.
The total number of MNIST data used in this experiment was 10,000, with 1,000 MNIST data for each label.
Figure~\ref{fig:mnist_dataset} illustrates an example of the dataset used in this experiment.

% \begin{figure}[tb!p]
%     \centering
%     \includegraphics[width=0.5\linewidth]{img/mnist_dataset.pdf}
%     \caption{Example of the dataset used in this experiment. Agent A uses unprocessed MNIST data as image observations, and Agent B uses MNIST data rotated 45° to the left as image observations.}
%     \label{fig:mnist_dataset}
% \end{figure}

\begin{table}[bt!p]
 % \begin{minipage}{.66\textwidth}
        \centering 
        \caption{Experimental results for MNIST data:
        The means $\pm$ standard deviations of ARI and kappa coefficient $\kappa$ of $10$ trials of are shown. The highest scores are in bold, and the second-highest scores are underlined.}
        \label{tab:result-mnist}
        %\scalebox{0.8}{
        \begin{tabular}{ccccc}
        \toprule
        \shortstack{Condition} & \shortstack{MI} & \multicolumn{1}{c}{\shortstack{ARI\\(Agent A)}} &\shortstack{ARI\\(Agent B)} & \multicolumn{1}{c}{$\kappa$} \\ \hline 
        MH naming game                  & \checkmark & \multicolumn{1}{c}{{\bf {0.78$\pm$0.04}}} & {\bf {0.78$\pm$0.03}} & \multicolumn{1}{c}{{\bf {0.91$\pm$0.02}}} \\ %\hline
        MH naming game                   &  & \multicolumn{1}{c}{\underline{0.71$\pm$0.04}} & \underline{0.72$\pm$0.03} & \multicolumn{1}{c}{\bf {0.91$\pm$0.03}} \\% \hline
        No communication              & \checkmark & \multicolumn{1}{c}{$0.65\pm0.04$} & $0.68\pm0.05$ & \multicolumn{1}{c}{$0.04\pm0.04$} \\% \hline
        No communication               &  & \multicolumn{1}{c}{$0.60\pm0.02$} & $0.64\pm0.03$ & \multicolumn{1}{c}{$0.01\pm0.05$} \\ %\hline
        All acceptance        & \checkmark & \multicolumn{1}{c}{$0.68\pm0.04$} & $0.65\pm0.03$ & \multicolumn{1}{c}{$0.81\pm0.03$} \\ %\hline
        All acceptance         &  & \multicolumn{1}{c}{$0.61\pm0.03$} & $0.63\pm0.05$ & \multicolumn{1}{c}{$0.83\pm0.04$} \\ \hline
     %   \hline
        Gibbs sampling   & \checkmark & \multicolumn{2}{c}{$0.81\pm0.03$} & -- \\%\hline 
        Gibbs sampling   &  & \multicolumn{2}{c}{$0.73\pm0.04$} & -- \\ \bottomrule
        \end{tabular}%}
%  \end{minipage}
%\end{tabular}\vspace{-5mm}
\end{table}

{\bf Compared method}: The proposed model, {\it MH naming game} (proposal), was assessed by comparing two baseline models and a topline model. In {\it No communication} (baseline 1), two agents independently form internal representations $z$ and sign $w$. No communication occurs between the two agents. In other words, the {\it No communication} model assumes two GMM+VAEs for Agents A and B and independently infers signs $w^A_d$ and $w^B_d$, respectively. {\it All acceptance} (baseline 2) is the same as the MH naming game, whose acceptance ratio is always $r = 1$ in MH communication (MH-COM in Algorithm~\ref{alg:MHNG}). Each agent always believes that the sign of the other is correct. In {\it Gibbs sampling} (topline), sign $w_d$ is sampled using the Gibbs sampler. This process directly uses $z^A_d$ and $z_d^B$, although no one can simultaneously examine the internal (i.e., brain) states of the two agents in human semiotic communication. As a result, the condition is not a model of emergent communication; instead, it is a topline as an inter-GMM+VAE centralized inference procedure.

{\bf Network architecture}: Convolutional and deconvolutional neural networks were simply employed for an encoder and a decoder of VAE. Appendix~\ref{apdx:architecture} presents the details.

{\bf Hyperparameters}: 
The hyperparameters of inter-GMM+VAE were set to $\alpha = 1.0$, $m = 0$, $\beta = 0.05I$, and $\nu = 12$.
The total number of signs was set to $K = 10$. The number of iterations of the MH naming game was $T = 100$.
The dimension of the latent variables $z^*_d$ was set to $12$, and the number of the training iterations of VAE for each update was set to $100$.
Adam, with a learning rate of $0.001$, was used as an optimizer.
The MI of VAE and GMM was conducted five times.

{\bf Evaluation criteria}: ARI~\cite{hubert1985comparing} was used to evaluate the unsupervised categorization performance of each agent through the MH naming game.
An ARI close to 1 indicates high categorization performance, whereas an ARI close to 0 indicates low performance. In contrast to the precision calculated by comparing the estimated labels and ground truth labels, ARI can consider label-switching effects in clustering.
The kappa coefficient $\kappa$ assessed the degree to which the two agents shared signs~\cite{cohen1960coefficient}.
For more details, please refer to Appendix~\ref{apdx:ari}.

{\bf Other conditions}: Experiments 1 and 2 were conducted using an Intel Core i9-9900K CPU with 1 $\times$ NVIDIA GeForce RTX2080 8GB GDDR6.

\subsection{Result}
\begin{figure}[tb!p]
    \centering
    \includegraphics[width=1.0\linewidth]{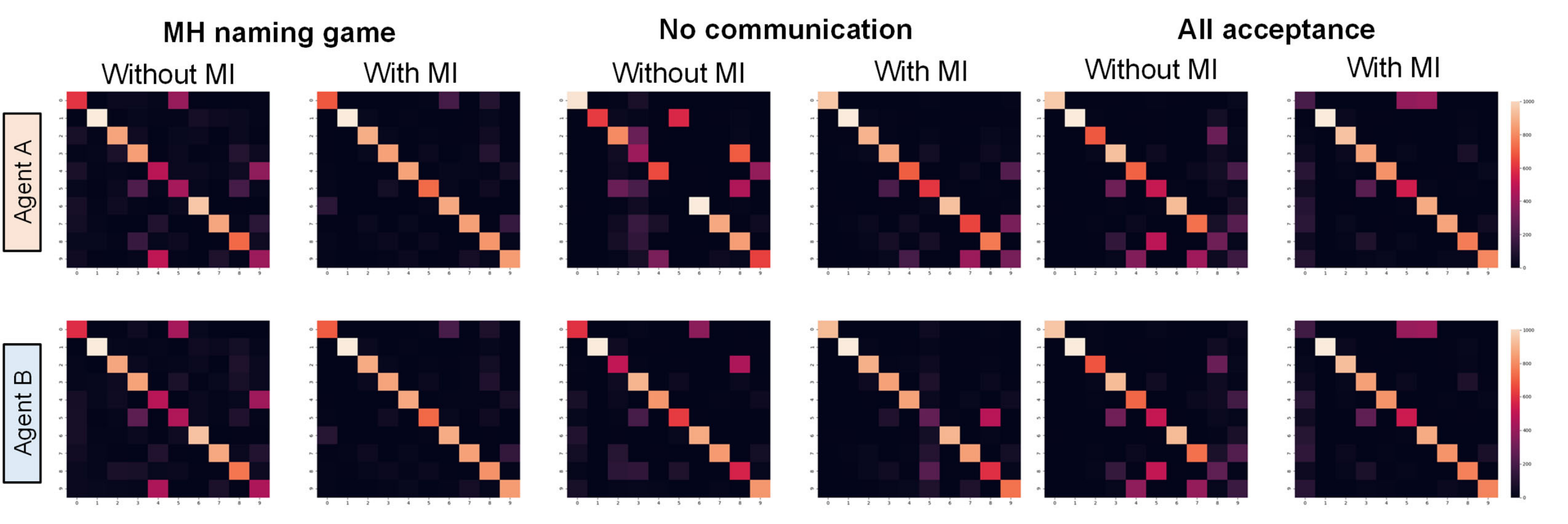}
    \caption{Confusion matrices in Experiment 1}
    \label{fig:confusion_matrix}
\end{figure}

{\bf Categorization and sharing signs}: Table~\ref{tab:result-mnist} presents the results of the ARI and the kappa coefficient values for each condition on the MNIST data.
Figure~\ref{fig:confusion_matrix} illustrates the confusion matrices of $w^A$ and $w^B$ for each condition.
The vertical axis represents the ground truth indices, and the horizontal axis represents the estimated signs, which are ordered for viewing.
The results demonstrate that the {\it MH naming game} leads two agents to categorize the objects at nearly the same level as the Gibbs sampling (topline), which is a centralized inference procedure\footnote{The performance seemingly remains relatively lower than the existing emergent communication models based on deep reinforcement learning. Notably, however, they use explicit feedback and negative samples to improve discriminability (i.e., discriminative models). We consider that their assumption is not natural from the developmental perspective, although specific performance measures could be improved. Therefore, the direct comparison between their and our assumptions is not straightforward.}. Additionally, symbols emerged and were used between the two agents. Interestingly, the MH naming game between two agents improved categorization without any additional supervision compared to the {\it no communication} conditions (i.e., perceptual categorization conducted by a single agent). This finding is regarded as an advantage of multimodal unsupervised learning. Inter-GMM+VAE is a multimodal extension of GMM+VAE as an integrative model, and the MH naming game is an approximated MCMC inference process. As a result, the MH naming game evidently utilizes various observations gathered from different agents to increase classification performance through the inference of $P(w_d|o^A_d, o^B_d)$.
{\it No communication}, certainly, could not share signs and exhibited a worse categorization performance than the MH naming game. {\it All acceptance} could share signs to a certain extent. Although {\it all acceptance} attempts to make each agent mimic the use of signs of the other, the procedure did not result in their sharing of signs at the same level as the MH naming game. The reason is that each agent in {\it all acceptance} must accept signs produced by the other, whose categorization may be immature or even incorrect.
As Figure~\ref{fig:confusion_matrix} suggests, communication in the {\it all acceptance} condition did not address the confusion between categories 0 and 5, whereas the MH naming game could.
In terms of the MI of GMM and VAE, MI enhanced classification performance in each condition.

\begin{figure}[tb!p]
%\begin{tabular}{cc}
%  \begin{minipage}{.39\textwidth}
    \centering
    \includegraphics[width=0.6\linewidth]{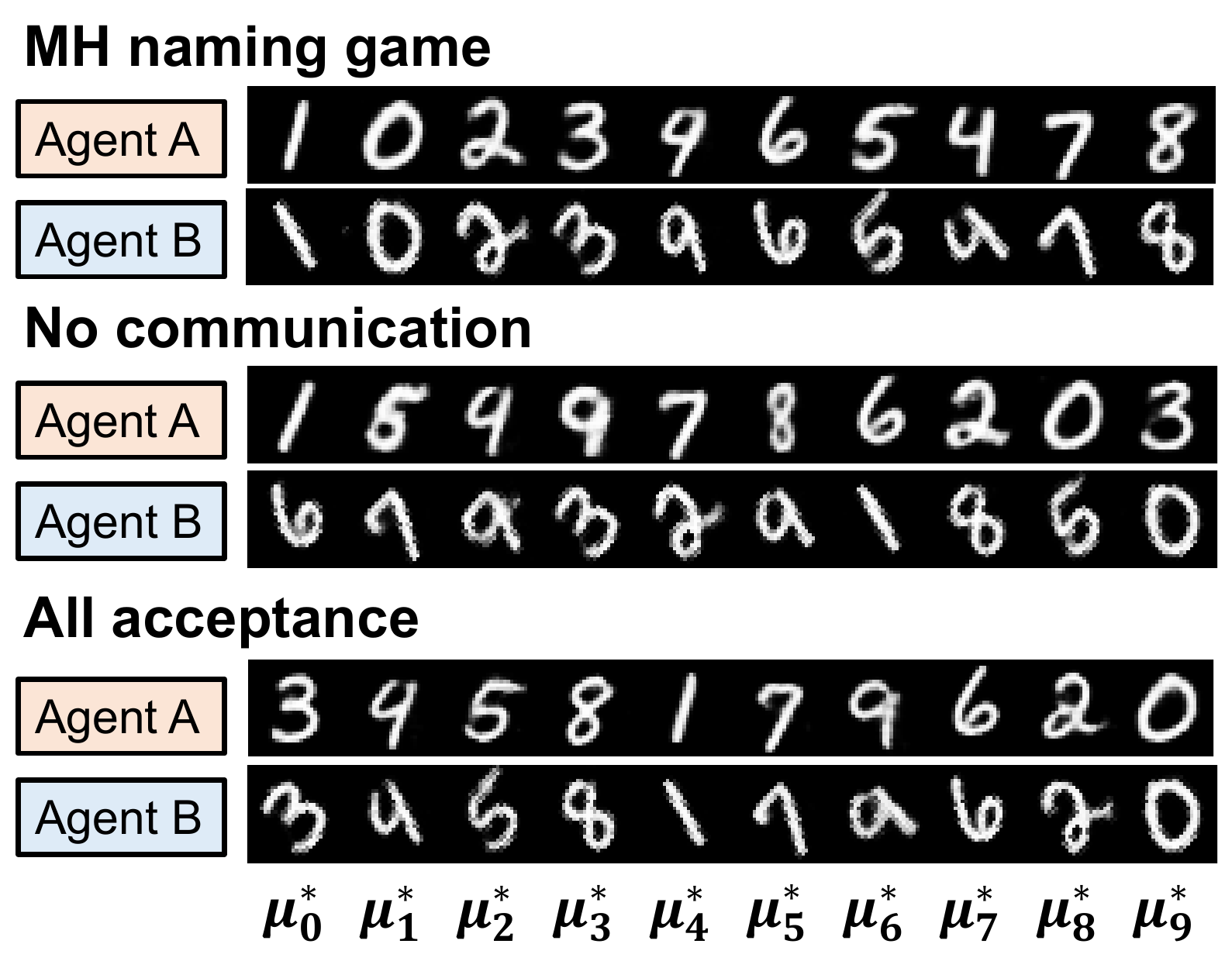}
    \caption{MNIST images recalled by each agent by signs}
    \label{fig:imagined_mnist}
%  \end{minipage}
\end{figure}
\begin{figure}[tb!p]
%  \begin{minipage}{.60\textwidth}
    \centering
    \includegraphics[width=0.7\linewidth]{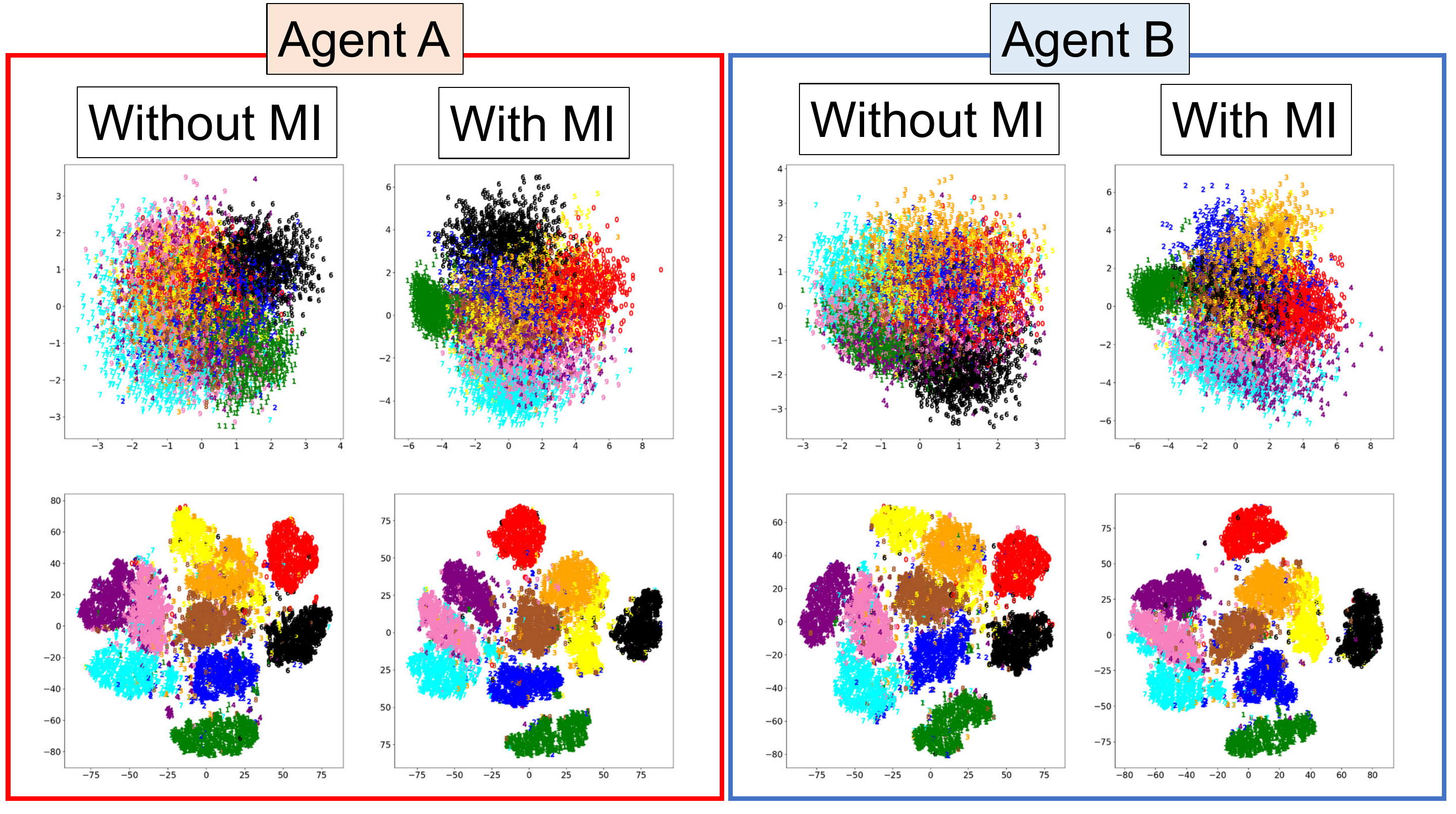}
    \caption{Visualization of the internal representations $z$ for MNIST data with (top) PCA and (bottom) t-SNE.}% Each color corresponds to each digit.}
    \label{fig:represetation_mnist}
%  \end{minipage}
%\end{tabular}\vspace{-5mm}
\end{figure}

{\bf Imagination from signs}:
Figure~\ref{fig:imagined_mnist} reveals images recalled from each emerged sign by each agent. The images corresponding to the sign $w$ were recalled by reconstructing observation $o$ from the mean vector of the $w$-th Gaussian distribution $\mu^*_w$.
% \begin{figure}[b]
%     \centering
%     \includegraphics[width=0.5\linewidth]{img/imagined_mnist.pdf}
%     \caption{MNIST images recalled by each agent by signs}
%     \label{fig:imagined_mnist}
% \end{figure}
In the MH naming game, each agent successfully reconstructed each number. Different digits from the same sign $w$ were rebuilt by agents in {\it no communication}. In {\it all acceptance}, the agents could nearly imagine digits from signs. However, digits 4 and 9 led to slight confusion, which corresponds to labels 0 and 5, respectively in Figure~\ref{fig:confusion_matrix} due to sorting.

{\bf Formation of internal representations}:
Figure~\ref{fig:represetation_mnist} illustrates the latent variables $z^A$ and $z^B$ of VAE (i.e., internal representations of each agent). GMM and VAE with and without MI are shown to demonstrate the effect of MI. For visualization, the study employed principal component analysis (PCA) and t-SNE~\cite{van2008visualizing}. The same color indicates the same digit.
The findings of visualization indicate that MI in the VAE+GMM and MH naming game (i.e., MI across two VAE+GMM), brought internal representations that correspond to the same digit closer together as a result of the prior distribution of the GMM.

% \begin{figure}[b]
%     \centering
%     \includegraphics[width=0.5\linewidth]{img/represetation_mnist.pdf}
%     \caption{Visualization of internal representations $z$ for MNIST data. Each color corresponds to each digit.}
%     \label{fig:represetation_mnist}
% \end{figure}

\section{Experiment 2: Fruits 360}\label{sec:fruit360}
\subsection{Conditions}
\begin{figure}[tb!p]
%\begin{tabular}{cc}
%  \begin{minipage}{.33\textwidth}
	\centering
    \includegraphics[width=0.5\linewidth]{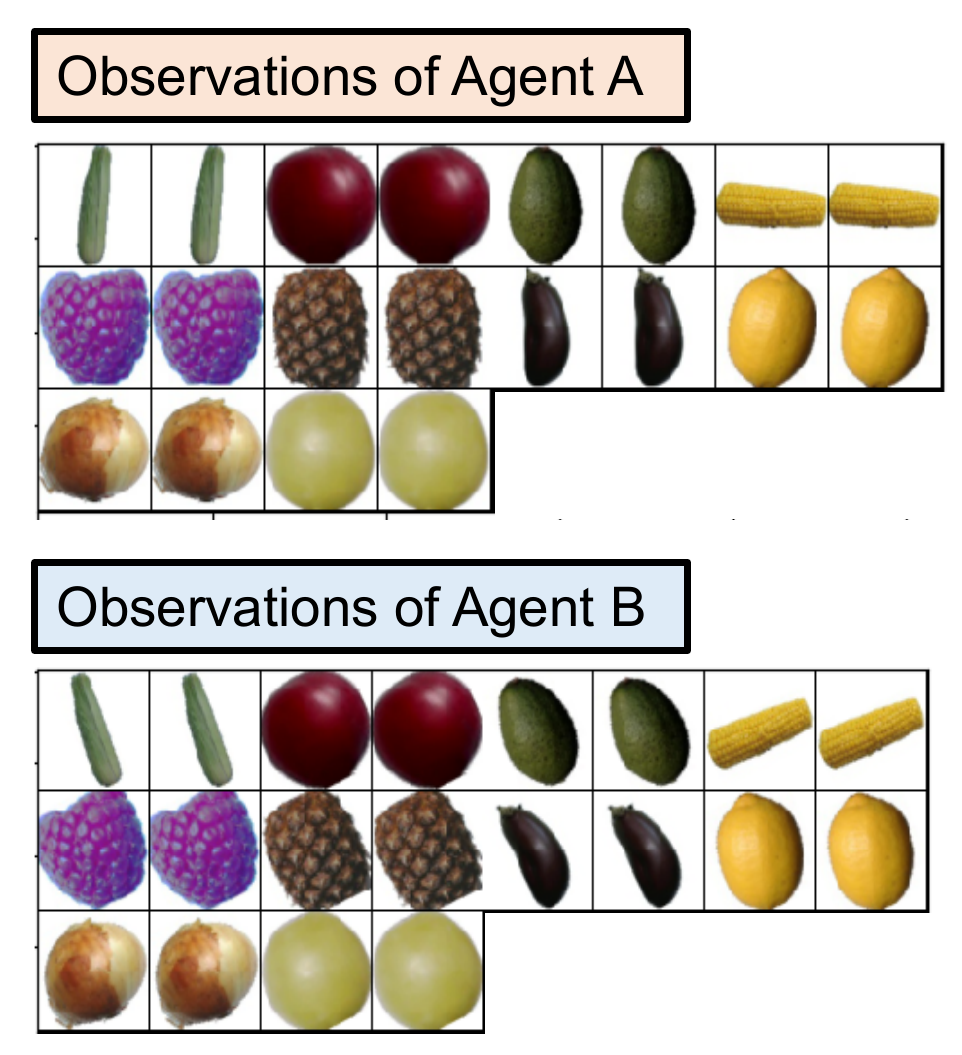}
    \caption{Fruits 360 dataset}
    \label{fig:fruit_dataset}
%  \end{minipage}
\end{figure}

{\bf Dataset}: To verify the proposed method on natural images, we used the Fruits 360 dataset\footnote{Fruits 360: \url{https://www.kaggle.com/moltean/fruits}}.
The Fruits 360 dataset consists of images of fruits and vegetables under a total of $131$ categories with RGB channels and $100 \times 100$ pixels.
We utilized raw Fruits 360 data for the observations of Agent A, and Fruits 360 data rotated 25° to the left for observations of Agent B, as in Experiment 1.
This model assumes that the two agents were looking at the same objects from different viewpoints.
This experiment employed a total of 2,350 Fruits 360 data points, with 235 Fruits 360 images used for each label.
In this experiment, the study used 10 out of 131 categories (i.e., Corn Husk, Cherry Wax Red, Avocado, Corn, Raspberry, Pineapple, Eggplant, Lemon, Onion White, and Grape White 2). Figure~\ref{fig:fruit_dataset} depicts the examples of the dataset used in this experiment.

\begin{table}[bt!p]
%  \begin{minipage}{.66\textwidth}
        \centering 
        \caption{Experimental results for Fruits 360 data: The means $\pm$ standard deviations of ARI and kappa coefficient $\kappa$ of $10$ trials are shown. The highest scores are in bold, and the second-highest scores are underlined.}
        \label{tab:result-fruit}
    %    \scalebox{0.75}{
        \begin{tabular}{ccccc}
        \toprule
        \shortstack{Condition} & \shortstack{MI} & \multicolumn{1}{c}{\shortstack{ARI\\(Agent A)}} &\shortstack{ARI\\(Agent B)} & \multicolumn{1}{c}{$\kappa$ } \\ \hline 
         MH naming game    & \checkmark & \multicolumn{1}{c}{{ $\mathbf{0.76 \pm 0.04}$}} & {{\bf $\mathbf{0.77 \pm 0.02}$}} & \multicolumn{1}{c}{\underline{$0.92 \pm 0.02$}} \\ %\hline
         MH naming game    &  & \multicolumn{1}{c}{\underline{$ 0.65\pm0.03$}} &{\underline{$ 0.65\pm0.03$}} & \multicolumn{1}{c}{{$\mathbf{0.94\pm0.02}$}} \\ %\hline
         No communication  & \checkmark & \multicolumn{1}{c}{$ 0.64\pm0.03$} &{\underline{$ 0.65\pm0.05$}} & \multicolumn{1}{c}{$ -0.08\pm0.05$} \\ %\hline
         No communication  &  & \multicolumn{1}{c}{$ 0.53\pm0.04$} & $ 0.56\pm0.02$ & \multicolumn{1}{c}{$ 0.01\pm0.06$} \\ %\hline
         All acceptance  & \checkmark & \multicolumn{1}{c}{$0.52\pm0.03$} & $0.50\pm0.03$ & \multicolumn{1}{c}{$0.58\pm0.04$} \\ %\hline
         All acceptance  &  & \multicolumn{1}{c}{$ 0.43\pm0.03$} & $ 0.43\pm0.02$ & \multicolumn{1}{c}{$ 0.42\pm0.03$} \\ %\hline
        Inter-DM~\cite{hagiwara2019symbol}        &  & \multicolumn{1}{c}{$0.59\pm0.04$} & $0.56\pm0.03$ & \multicolumn{1}{c}{$0.85\pm0.03$} \\ %\hline
        Inter-GMM       &   & \multicolumn{1}{c}{$ 0.02\pm0.07$} & $ 0.01\pm0.03$ & \multicolumn{1}{c}{$ 0.18\pm0.04$} \\ \hline%\hline
        Gibbs sampling  & \checkmark & \multicolumn{2}{c}{$ 0.74\pm0.05$} &    --       \\ 
        Gibbs sampling  &  & \multicolumn{2}{c}{$ 0.62\pm0.04$} & -- \\ \bottomrule                    
        \end{tabular}%}
%  \end{minipage}
%\end{tabular}
\end{table}

% \begin{figure}[tb!p]
%     \centering
%     \includegraphics[width=0.5\linewidth]{img/fruit_dataset.pdf}
%     \caption{Example of Fruits 360 dataset used in Experiment 2}
%     \label{fig:fruit_dataset}
% \end{figure}

{\bf Compared method}: In addition to the conditions used in Experiment 1, the study used inter-DM~\cite{hagiwara2019symbol} and inter-GMM to assess the contribution of representation learning by VAE. Inter-DM and inter-GMM categorize using the Dirichlet mixtures and Gaussian mixture models, respectively. Observations were represented as bag-of-features (BoF) representations for inter-DM and inter-GMM. The BoF representations were standardized for inter-GMM.

{\bf Network architecture}: Convolutional and deconvolutional neural networks were employed simply as the encoder and decoder of the VAE, respectively. Appendix~\ref{apdx:architecture} illustrates the details.

{\bf Hyperparameters} and {\bf evaluation criteria}: 
The same hyperparameters and evaluation criteria in Experiment 1 were utilized.

\subsection{Result}
{\bf Categorization and sharing signs}: Table~\ref{tab:result-fruit} presents the results of the ARI and kappa coefficient values for each condition on the Fruits 360 dataset. Among the compared approaches, the MH naming game with MI, which is the proposed method, notably marked the highest score in ARIs. Gibbs sampling, which is the proposed method, performed at the same level as the topline approach. According to the theory, the suggested approach and Gibbs sampling can sample from the same posterior distribution. This finding supports the theoretical implication (Theorem 1).

Considering $\kappa$, the MH naming game without and with MI took the first and second highest scores, respectively.
The flexibility of the encoder occasionally produces a shift in the connection between internal representations $z$ and signs $w$. This tendency may have rendered $kappa$ of the MH naming game with MI slightly worse.
In all aspects, inter-GMM+VAE exceeded inter-DM and inter-GMM. This result demonstrates that VAE representation learning may identify acceptable representations for emergent communication.

%\begin{wrapfigure}{r}[0pt]{0.5\linewidth}
\begin{figure}[tb!p]
    \centering
    \includegraphics[width=0.8\linewidth]{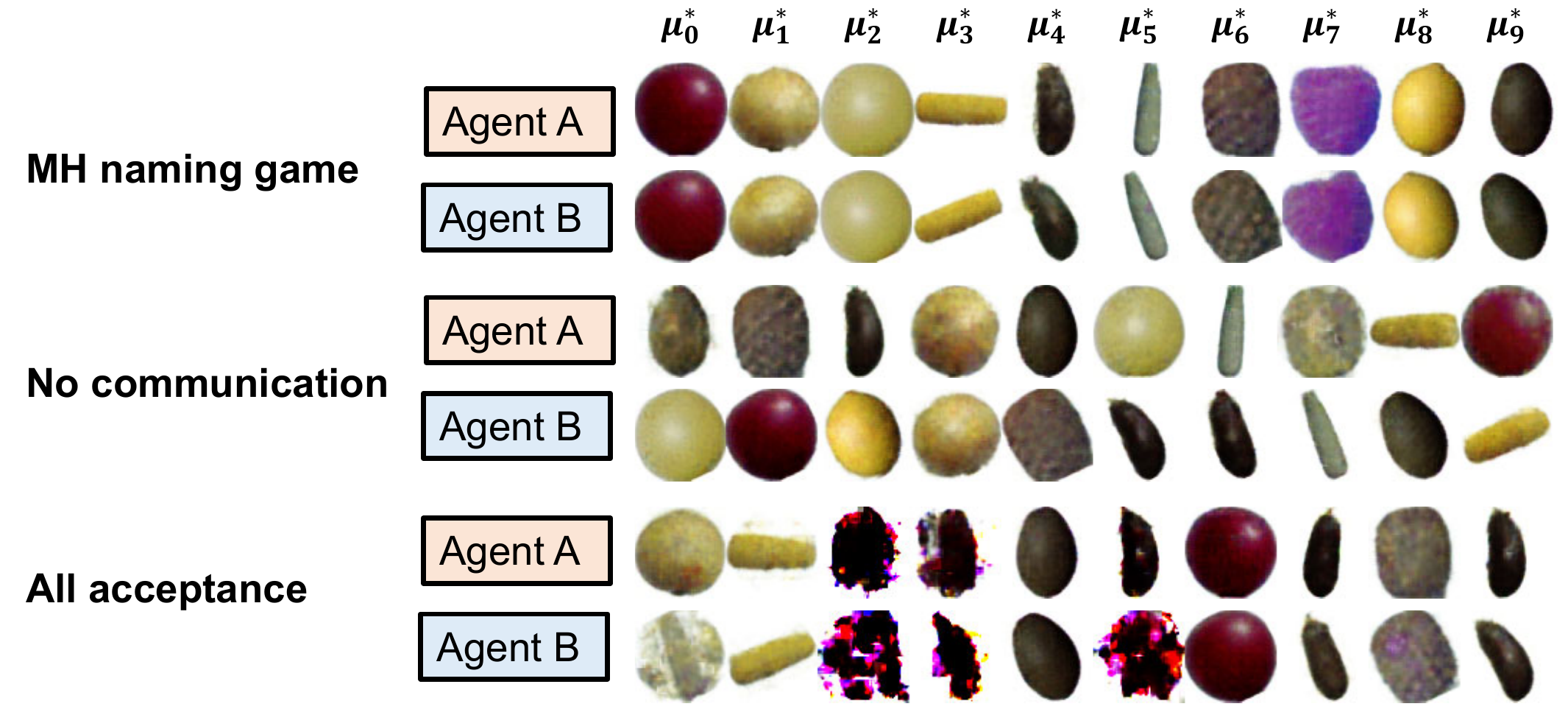}
    \caption{Fruits 360 images recalled by each agent by signs}
    \label{fig:imagined_fruit}
\end{figure}
%\end{wrapfigure}

{\bf Imagination from signs}:
Figure~\ref{fig:imagined_fruit} presents images recalled from each emerged sign by each agent.
In the MH naming game, each agent successfully recalled each fruit image. Alternatively, {\it all acceptance} recalls the same fruit for 7 and 9 and collapsed imagery for 2, 3, and 5. This result is due to the inability of the agent to appropriately create internal representations and fruit categories.

The results reveal that the MH naming game on inter-GMM+VAE enabled two agents to cooperatively create internal representations, execute categorization, and share signs via the decentralized Bayesian inference.

\section{Conclusion and discussion}\label{sec:conclusion}
This work detailed a new model for emergent communication based on a deep PGM. It defined the MH naming game was defined by generalizing prior works~\cite{hagiwara2019symbol,hagiwara2022multiagent} and demonstrated that the MH naming game is the same as a form of MH algorithm for a PGM, which is an integrative model that combines two agents performing representation learning and participating in the naming game.
From this viewpoint, symbol emergence and semiotic communication are regarded as {\it decentralized approximate Bayesian inference} and {\it inter-personal cross-modal inference}, respectively.
To achieve emergent communication and symbol emergence based on raw images, the study proposed a deep generative model called inter-GMM+VAE. An MH naming game between two GMM+VAEs and MI between GMM and VAE comprised the inference process.
Experiments using the MNIST and Fruits 360 datasets illustrated that the model enables two agents to simultaneously form internal representations and categories and share signs. Moreover, the study demonstrated that a listener could reconstruct appropriate images from the signs of a speaker.

{\bf Theoretical extensions}:~
The proposed generative model-based approach to emergent communication is relatively generic and leaves potential for future expansions. In inter-GMM+VAE, the sign $w$ is assumed to be a categorical variable (i.e., a discrete sign). However, the MH naming game itself does not restrict $w$ as a categorical variable. A conceivable path is extending $w$ to word sequences while considering compositionality. The number of sign types, which correspond to Gaussian components, is fixed in inter-GMM+VAE. To render it flexible, using Bayesian nonparametrics (e.g., Dirichlet process GMM) is a possible solution~\cite{teh2006hierarchical,blei2006variational,nakamura2011multimodal}.
In addition, the generative model for an agent can be replaced with other sophisticated models. The current study employed GMM+VAE for simplicity. It is known that a multinomial VAE performs object categorization.  For example, using a multimodal VAE instead of the unimodal VAE is one possible extension ~\cite{kuniyasurobot}. Another task is to investigate improved models and network architecture. Another problem is extending the MH naming game from a two-agent party to an N-agent game.

{\bf Collective predictive coding hypothesis}:~
One of the implications of the MH naming game is that if we, as humans, perform representation learning, name objects based on the perceptual status of each person, and determine whether or not we accept or reject the naming of another person in a probabilistic manner based on his/her belief, then we can collectively estimate categories of objects based on the distributed perceptions of people. In general, PGMs are trained to predict observations, and representations (i.e., latent variables) are encoded through inference. Such a process is called {\it predictive coding}, which is also a strong explanatory theory for the brain~\cite{hohwy2013predictive,FRISTON2021573}.
Based on these notions, we hypothesize that humans are collectively inferring latent representations to better represent the world by integrating partial sensory observations obtained by each agent. In other words, symbol systems, especially {\it language}, are formed through {\it collective predictive coding} in a symbol emergence system.
We may call this idea {\it collective predictive coding hypothesis}.

{\bf Society-wide free energy principle}:~
This term can be rephrased by another term called {\it free energy principle}. The inference of latent variables (i.e., representations) is formulated with free-energy minimization from the viewpoint of variational inference~\cite{FRISTON2021573}. The free energy principle is a general notion of predictive coding and an influential idea in neuroscience. Scholars frequently mention that the human brain performs free-energy minimization. Beyond the individual free energy principle, the collective predictive coding hypothesis suggests that the human society performs free-energy minimization at the societal level by making symbol systems emerge. This speculation introduces the idea that symbol emergence is driven by the {\it society-wide free energy principle}.

{\bf Social representation learning}:~
From the viewpoint of a deep generative model, the performance of the agents in the experiments was only representation learning based on multimodal information.
Nevertheless, the agents conducted representation learning in which representations are not only organized inside the brain but also formed as a symbol system at the societal level. In this context, we can call the symbol emergence {\it social representation learning}.

We can speculate that the ability of individuals to accept or reject the utterances of others may be a basic genesis of human language.
To investigate this viewpoint, we are also investigating whether or not human participants follow the likelihood of acceptance when playing a similar naming game. Exploring and testing the collective predictive coding hypothesis is also a future challenge.

\bibliographystyle{tADR}
%\bibliography{tADR}
%\bibliographystyle{plain}
\bibliography{articles,cpc}

\section*{Acknowledgments}
This study was partially supported by the Japan Society for the Promotion of Science (JSPS) KAKENHI under Grant JP21H04904 and JP18K18134 and by MEXT Grant-in-Aid for Scientific Research on Innovative Areas 4903 (Co-creative Language Evolution), 17H06383.

%\label{lastpage}

%\clearpage

%\begin{comment}

\appendix

%\appendix
\section{Appendix}
\subsection{Acceptance rate in the MH naming game}\label{apdx:proof}
Given $P(z)$ and $Q(z^\star|z)$ are target and proposal distributions, respectively, where $z^\star$ is a proposed sample, then the acceptance rate of MH algorithm is as follows:
\begin{align}
R(z,z^\star)&={\rm min}(1,\bar{R}(z,z^\star))\\
\bar{R}(z,z^\star)&=\frac{P(z^\star)Q(z|z^\star)}{P(z)Q(z^\star|z)} .
\end{align}
In the MH naming game, the target distribution for $w_d$ is $P(w_d|z_d^{Sp}, z_d^{Li}, \phi^{Sp}, \phi^{Li})$ and the proposal distribution is $P(w_d|z_d^{Sp}, \phi^{Sp})$.
\begin{align}
&\bar{R}(w^{Li},w^{Sp})\\
&= \frac{P(w^{Sp}|z^{Sp},z^{Li}, \phi^{Sp}, \phi^{Li})P(w^{Li}|z^{Sp}, \phi^{Sp})}
                   {P(w^{Li}|z^{Sp},z^{Li}, \phi^{Sp}, \phi^{Li})P(w^{Sp}|z^{Sp}, \phi^{Sp})} \\
                  &=  \frac{\left[P(z^{Li}, \phi^{Li}|w^{Sp},z^{Sp},\phi^{Sp})P(w^{Sp}|z^{Sp}, \phi^{Sp})/P(z^{Li}, \phi^{Li}|z^{Sp}, \phi^{Sp})\right]P(w^{Li}|z^{Sp},\phi^{Sp} )}
                  {\left[P(z^{Li}, \phi^{Li}|w^{Li},z^{Sp},\phi^{Sp})P(w^{Li}|z^{Sp}, \phi^{Sp})/P(z^{Li}, \phi^{Li}|z^{Sp}, \phi^{Sp})\right]P(w^{Sp}|z^{Sp},\phi^{Sp} )} \\    
                    &=   \frac{P(z^{Li}, \phi^{Li}|w^{Sp},z^{Sp},\phi^{Sp})}
                    {P(z^{Li}, \phi^{Li}|w^{Li},z^{Sp},\phi^{Sp})}\\
                     &=   \frac{P(z^{Li}| \phi^{Li}, w^{Sp},z^{Sp},\phi^{Sp}) P(\phi^{Li}| w^{Sp},z^{Sp},\phi^{Sp})      }
                    {P(z^{Li}| \phi^{Li}, w^{Li},z^{Sp},\phi^{Sp})P(\phi^{Li}|w^{Li},z^{Sp},\phi^{Sp})}\\     
                     &=   \frac{P(z^{Li}| \phi^{Li}, w^{Sp})     }
                    {P(z^{Li}| \phi^{Li}, w^{Li})} 
\end{align}
As a result, the acceptance rate in the MH naming game satisfies the condition of the MH algorithm.

\newpage
\subsection{Overall procedure of the MH naming game for inter-GMM+VAE}\label{apdx:procedure}
\begin{figure}[h]
    \centering
    \includegraphics[width=1.0\linewidth]{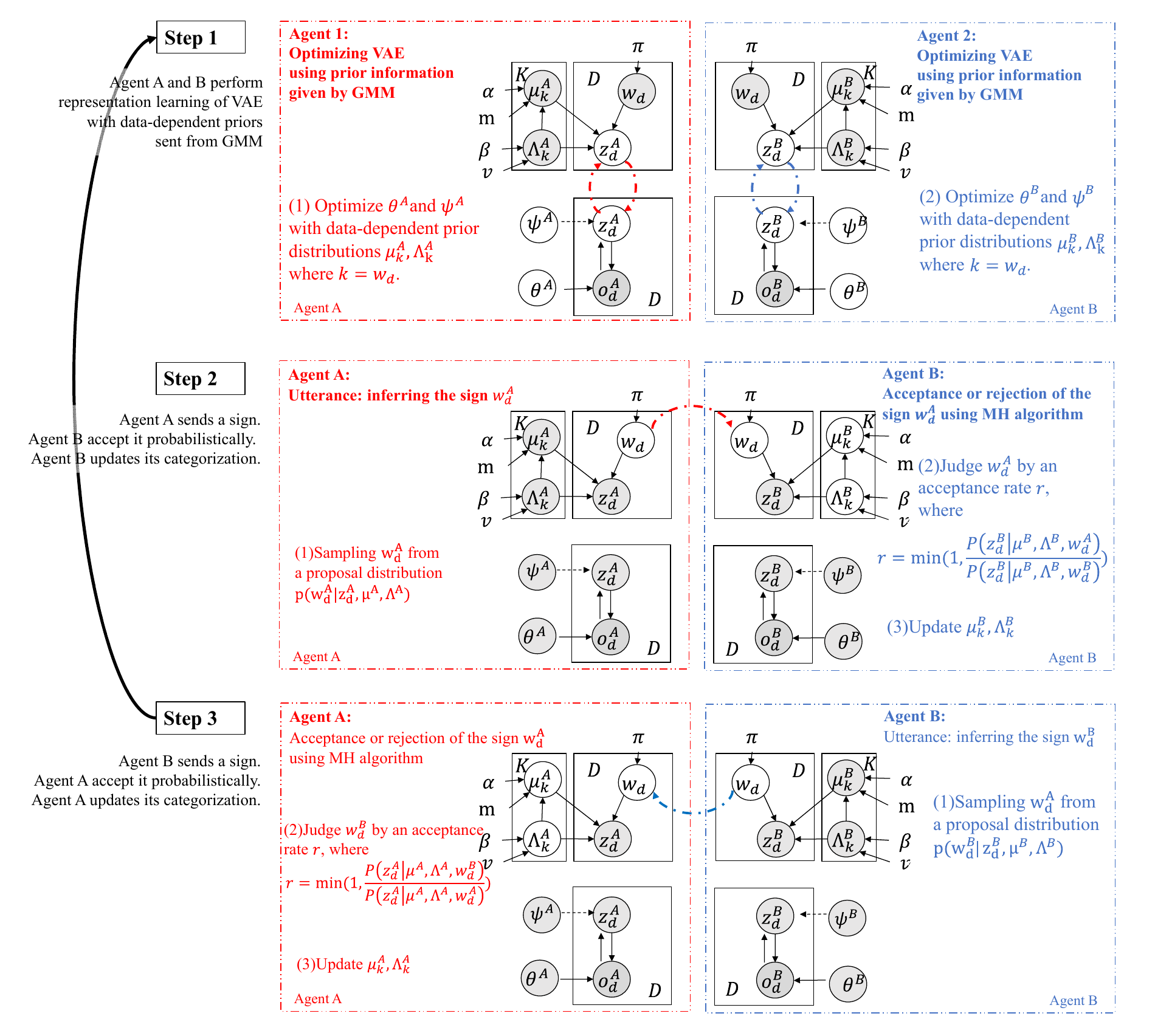}
    \caption{Overall procedure of MH naming game for Inter-GMM+VAE}
    \label{fig:procedure}
\end{figure}

Figure~\ref{fig:procedure} presents a diagram depicting the overall MH naming game (i.e., inference procedure).
In Figure~\ref{fig:procedure}, $\psi^*$ denotes the parameter of the inference network.

\newpage
\subsection{Network architecture}\label{apdx:architecture}
\begin{figure}[tbh!p]
    \centering
    \includegraphics[width=0.5\linewidth]{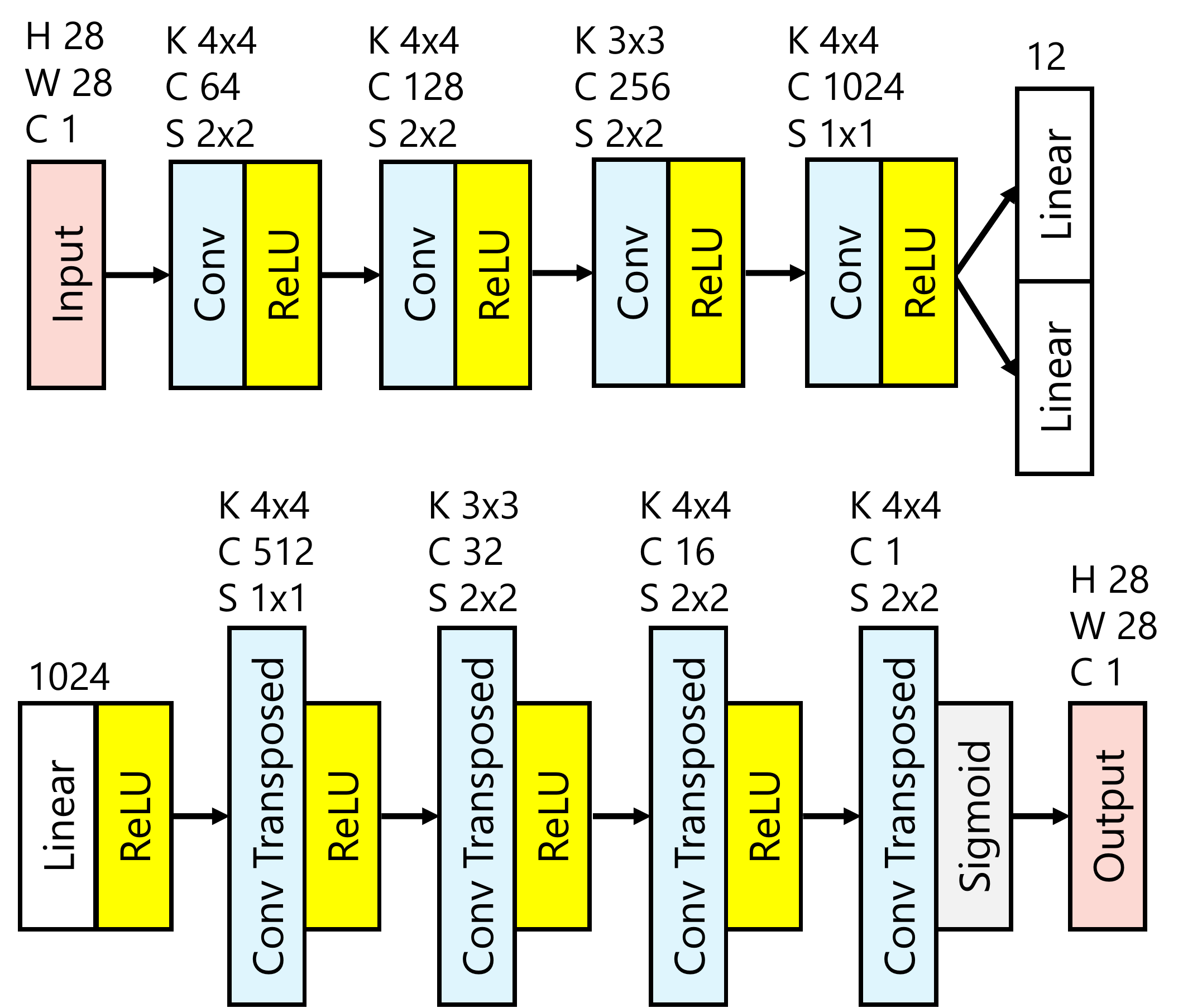}
    \caption{Network architecture of VAE in Experiment 1}
    \label{fig:network_mnist}
% \end{figure}
% \begin{figure}[h]
    \centering
    \includegraphics[width=0.5\linewidth]{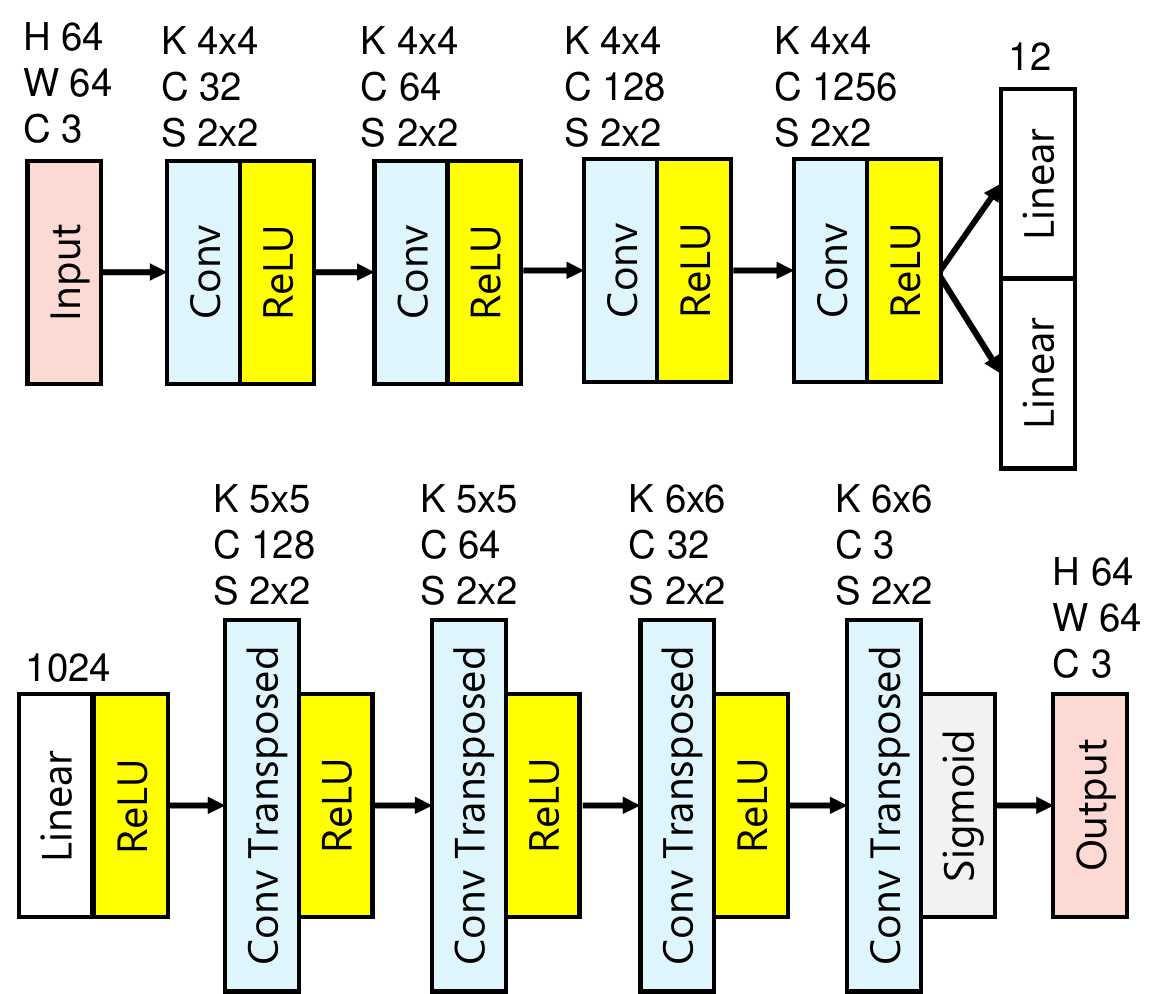}
    \caption{The network structure of VAE used in this Experiment 2}
    \label{fig:network_fruit}
\end{figure}

Figure~\ref{fig:network_mnist} displays the network structure of VAE used in Experiment 1. $H$ and $W$ represent the height and width of the image, respectively; $K$ denotes the number of kernels, $C$ pertains to the number of filters in the output, and $S$ stands for the number of strides. 
The layers ${\mathrm Conv}$, ${\mathrm Conv\ Transposed}$, $\mathrm Linear$, ${\mathrm ReLU}$ and ${\mathrm Sigmoid}$ denote the convolution, transposed convolution, and fully-connected layers, the ReLU function, and the Sigmoid function, respectively.
Figure~\ref{fig:network_fruit} illustrates the network structure of VAE used in Experiment 2 in the same manner.

\newpage
\subsection{ARI and kappa coefficient}\label{apdx:ari}
ARI is extensively used to evaluate clustering performance by comparing clustering results with ground truth labels.
In contrast to the precision calculated by comparing the estimated labels and ground truth labels for evaluating the classification systems trained by supervised learning, ARI can consider label switching effects in clustering.
ARI is defined by the following equation (\ref{eq:ari}). Notably, ${\rm RI}$ denotes ${\rm Rand Index}$.
\begin{align}
{\rm ARI} = \frac{\rm{RI - Expected\ RI}}{{\rm Max\ RI - Expected\ RI}}\label{eq:ari}
\end{align}
For more details, please refer to \cite{hubert1985comparing}.

The kappa coefficient $\kappa$ is defined by the following equation (\ref{eq:kappa}):
\begin{align}
\kappa = \frac{C_o - C_e}{1-C_e} \label{eq:kappa}
\end{align}
where $C_o$ denotes the degree of agreement of signs across agents, and $C_e$ represents the expected value of the coincidental sign agreement.
The evaluation criteria of $\kappa$ are as follows~\cite{landis1977measurement}:
\begin{enumerate}
    \item Almost perfect agreement: ($1.0 \geq \kappa > 0.80$)
    \item Substantial agreement: ($0.80 \geq \kappa > 0.60$)
    \item Moderate agreement: ($0.60 \geq \kappa > 0.40$)
    \item Fair agreement: ($0.40 \geq \kappa > 0.20$)
    \item Slight agreement: ($0.20 \geq \kappa > 0.00$)
    \item No agreement: ($0.0 \geq \kappa$)
\end{enumerate}

\end{document}